\documentclass[runningheads]{llncs}

\usepackage{eccv}

\usepackage{eccvabbrv}

\usepackage{graphicx}
\usepackage{booktabs}
\usepackage{multirow}
\usepackage{float}
\usepackage{fontawesome5}

\usepackage[accsupp]{axessibility}

\usepackage{hyperref}
\hypersetup{hidelinks}  
\usepackage{orcidlink}

\definecolor{revblue}{RGB}{0,64,196}
\newcommand{\rev}[1]{#1}
\begin{document}
\title{MEVL-STP: Multi-Encoder and Vision Language Model for Arbitrarily Shaped Scene Text Spotting}
\titlerunning{Multi-Encoder and VLM Recognition Text Spotting }

\author{Aman Anand\inst{1} \and
Partha Pratim Roy\inst{2} \and
Palaiahnakote Shivakumara\inst{3}}

\authorrunning{A. Anand, P. P. Roy, P. Shivakumara}

\institute{
Rajiv Gandhi Institute of Petroleum Technology, India\\
\email{amananand7080@gmail.com}
\and
Indian Institute of Technology (ISM) Dhanbad, India\\
\email{parthapratim@iitism.ac.in}
\and
University of Salford, UK\\
\email{s.palaiahnakote@salford.ac.uk}
}

\maketitle

\begin{abstract}
Scene text spotting remains challenging for arbitrarily shaped text instances such as curved signs and dense multi-oriented characters in natural images, where tightly coupled architectures propagate localization errors directly into recognition failures. We present a two-stage pipeline that combines multi-encoder segmentation with vision-language model recognition to address this problem. In the detection stage, six frozen vision encoders (CLIP, DINOv2, SigLIP, EVA-CLIP, SAM, and ConvNeXt) extract complementary features spanning semantic, spatial, and texture spectra, which are fused through a trainable hierarchical Feature Pyramid Network with channel attention and decoded via a deep-supervision Progressive Scale Expansion network to generate precise instance-level text masks. By keeping the encoders frozen, their independently learned feature spaces remain orthogonal during fusion, preventing the feature homogenization that degrades boundary precision in single-backbone detectors. The detection stage produces tight polygon masks that conform to the actual shape of curved and arbitrarily oriented text, rather than axis-aligned rectangles that inevitably include background content. In the recognition stage, these polygon-masked crops isolate the target text from surrounding clutter, allowing a Qwen3-VL-8B-Instruct model, fine-tuned via Low-Rank Adaptation on polygon-cropped scene text, to focus purely on reading the text without interference from neighbouring words or background noise. Without any synthetic pretraining data, our method achieves 91.99\% detection F-measure and 85.86\% end-to-end H-mean on CTW1500, setting a new state of the art and achieving strong performance on Total-Text and ICDAR 2015 without any synthetic training data. Code is available at \href{https://github.com/doubleblind-afk/MEVL-STP}{\faGithub\ \textnormal{MEVL-STP}}.

\end{abstract}

\section{Introduction}

\begin{figure*}[!htb]
    \centering
    \includegraphics[width=0.88\textwidth]{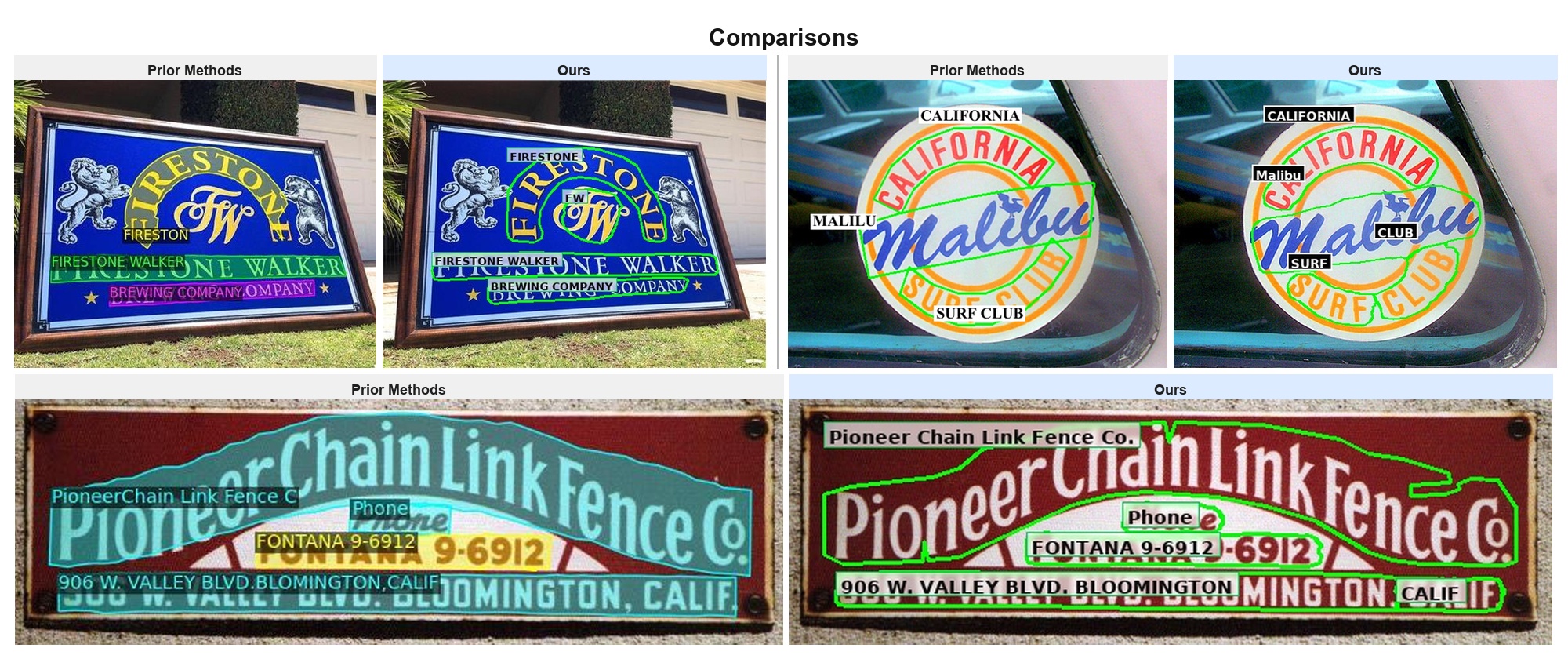}
   \caption{Qualitative comparison on challenging CTW1500 test samples. Each pair shows DeepSolo~\cite{ye2023deepsolo} (left) versus our method (right). \rev{In each left image, DeepSolo truncates the curved instance, breaks a single word into fragments, or misreads characters, which the broken and partial polygons and the wrong transcriptions make visible. In the right image of the same pair, our polygons stay tight, follow the full character baseline, and the VLM reads the text correctly. In short, our method finds the full text and reads it correctly.}}

    \label{fig:sample_comparison}
\end{figure*}

Text spotting, the joint task of detecting and recognizing text in natural images, has a wide range of real world applications, including autonomous driving, augmented reality, document understanding, and multilingual translation services~\cite{bai2016detecting,long2021scene}. Despite recent advances in transformer-based architectures~\cite{ye2023deepsolo,zhang2022testr}, accurately spotting arbitrarily shaped scene text remains challenging. Two particular difficulties stand out: curved text, where character baselines follow non-linear paths along circular signs or arched banners, and dense multi-oriented text, where closely spaced instances at varying angles cause detection overlap and recognition interference. As illustrated in Fig.~\ref{fig:sample_comparison}, existing methods frequently produce incomplete detections or misread such instances.

Unified end-to-end architectures jointly optimize detection and recognition through a shared backbone~\cite{ye2023deepsolo,pang2024tts,zhang2022testr,ye2024deepsolopp}. By formulating text spotting as a set prediction problem, these methods eliminate many hand crafted components. However, the shared representation creates a tight coupling between localization and recognition, where errors in one task directly degrade the other, making it difficult to optimize each task independently~\cite{xie2025dntextspotter}. To mitigate this coupling, two stage detect then recognize methods~\cite{wang2017crnn,qiao2020textperceptron,wang2018fots} decouple the tasks into a dedicated detector followed by a separate recognizer, enabling independent optimization. However, these methods typically rely on a single CNN or ViT backbone that may lack the rich semantic and structural cues needed for challenging curved and dense multi-oriented text instances~\cite{wang2019psenet,long2018textsnake,liao2020dbnet}.

A critical observation motivating our work is that detection accuracy directly bounds end-to-end performance. Prior studies~\cite{su2025lranet,huang2023estextspotter} demonstrate that inaccurate localization propagates cumulative errors that recognition modules cannot fully recover, particularly for curved text where tight polygon boundaries are essential to isolate target characters from adjacent instances. Meanwhile, the emergence of large scale vision language models (VLMs) has transformed text recognition. Models pre-trained on millions of image text pairs~\cite{qwen3technicalreport,radford2021clip} encode broad world knowledge and linguistic priors, making them naturally effective at reading multi-oriented, partially occluded, and context dependent scene text compared to traditional attention based decoders trained from scratch.

To address the challenges of curved and dense multi-oriented text spotting, we propose a decoupled framework that separates multi-encoder segmentation based detection from vision language model based recognition. In the detection stage, six frozen foundation encoders covering semantic, spatial, and texture feature spaces are fused through a trainable hierarchical Feature Pyramid Network with channel attention and decoded via deep-supervision Progressive Scale Expansion to produce precise polygon masks. In the recognition stage, these polygon crops isolate the target text from background clutter, enabling a Qwen3-VL model fine-tuned via Low-Rank Adaptation to read arbitrarily shaped text.

Our main contributions are as follows. First, we introduce a multi-encoder segmentation module that fuses six frozen foundation encoders spanning semantic, spatial, and texture feature spaces through a trainable hierarchical FPN with channel attention, capturing complementary visual cues for precise curved and dense multi-oriented text detection (91.99\% F1) without synthetic data. Second, we demonstrate that tight polygon masking from our detection stage isolates arbitrary text shapes from background noise, enabling a LoRA-adapted VLM to serve as a highly effective recognizer (94.23\% accuracy). Third, we achieve state-of-the-art end-to-end text spotting results on CTW1500 (85.86\% H-mean), validating that decoupled multi-encoder detection with VLM recognition effectively addresses arbitrarily shaped scene text.

\vspace{-2.2em}
\section{Related Work}

Existing scene text spotting methods can be broadly classified into single-stage end-to-end approaches and two-stage detection-recognition approaches.
\vspace{-0.1em}
\subsection{Single-Stage End-to-End Methods}

Early single-stage methods integrate detection and recognition within a shared backbone. OmniParser V2~\cite{yu2025omniparser} utilizes Structured Points of Thought prompts to jointly handle text spotting and layout analysis, while SemiETS~\cite{luo2025semiets} applies semi-supervised learning to enforce consistency between localization and transcription. However, these methods suffer from tight error coupling, where localization errors directly degrade recognition, making independent optimization difficult.

Transformer-based architectures address this limitation through set prediction mechanisms that improve detection flexibility. DNTextSpotter~\cite{xie2025dntextspotter} adopts denoising queries from DETR and represents arbitrary text using Bezier curves, though its iterative refinement increases training cost. SwinTextSpotter v2~\cite{swintextspotterv2} introduces a feature-alignment feedback loop between detection and recognition, adding architectural complexity. InstructOCR~\cite{instructocr2025} frames text spotting as an instruction-following task, providing format flexibility at the cost of heavy prompt engineering. While transformers improve detection, these single-stage methods still share a backbone across both tasks, limiting their ability to independently handle the challenges of curved and dense multi-oriented text.

Some methods specifically target curved text. IATS~\cite{zhang2024iast} models reading order via thin-plate spline transformations, but its specialized sampling struggles to generalize to standard horizontal text. CRENet~\cite{crenet2025} detects individual characters before grouping them into words, handling arbitrary shapes at the cost of grouping-stage errors and high model complexity. These methods highlight that curved text remains a persistent challenge for single-stage architectures.
\vspace{-0.1em}
\subsection{Two-Stage Detection and Recognition}

Two-stage approaches decouple detection and recognition into separate modules, enabling independent optimization of each task. Self-Calibrated Two-Stage Spotters~\cite{selfcalib2025} apply contrastive learning for detection representations alongside a transformer-based recognizer. TextBlockV2~\cite{lyu2024textblockv2} detects text blocks via segmentation before applying pre-trained language models for recognition. While decoupling resolves error coupling, these methods still rely on a single backbone for detection, which may lack the diverse semantic and structural cues needed for curved and dense multi-oriented text.

Our approach also adopts a decoupled two-stage pipeline but differs in two key aspects. First, we fuse six frozen foundation encoders through a trainable hierarchical FPN with channel attention rather than relying on a single backbone, capturing complementary semantic, spatial, and texture features for detection without synthetic data. Second, we apply parameter-efficient LoRA fine-tuning to a vision-language model for recognition rather than training a recognizer from scratch. This combination directly addresses the two core challenges: the multi-encoder ensemble provides the diverse feature coverage needed for precise curved text boundaries, while the VLM leverages broad linguistic priors to handle dense multi-oriented text where adjacent instances cause recognition interference.

\vspace{-0.5em}
\section{Methodology}
\begin{figure*}[!htb]
\centering
\includegraphics[width=0.76\textwidth]{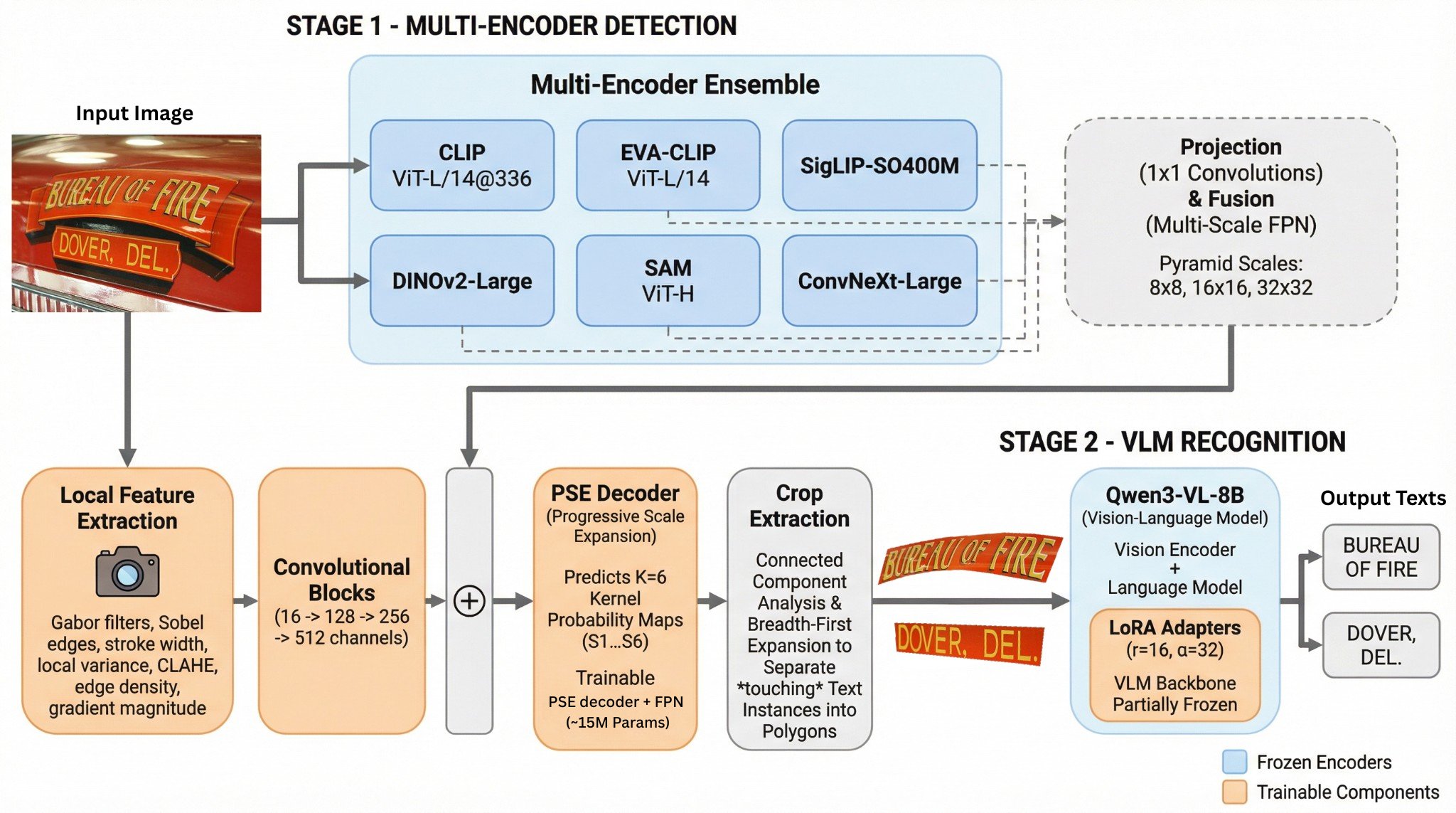}
\caption{Overview of our two-stage pipeline. Stage 1 extracts features from six frozen encoders and local features, followed by a PSE Decoder to perform text segmentation at instance level. In Stage 2, we use a fine-tuned Qwen3-VL-8B model along with LoRA adapters to perform text recognition from extracted crops.}
\label{fig:pipeline}
\end{figure*}

We propose a two-stage text spotting pipeline for arbitrarily-shaped scene text. The first stage employs a multi-encoder ensemble with Progressive Scale Expansion (PSE)~\cite{wang2019psenet} for instance-level text detection. The second stage utilizes a fine-tuned Vision-Language Model (VLM) for text recognition. Figure~\ref{fig:pipeline} illustrates the overall architecture.

\subsection{Multi-Encoder Feature Extraction}

\begin{figure}[!htb]
\centering
\includegraphics[width=0.62\linewidth]{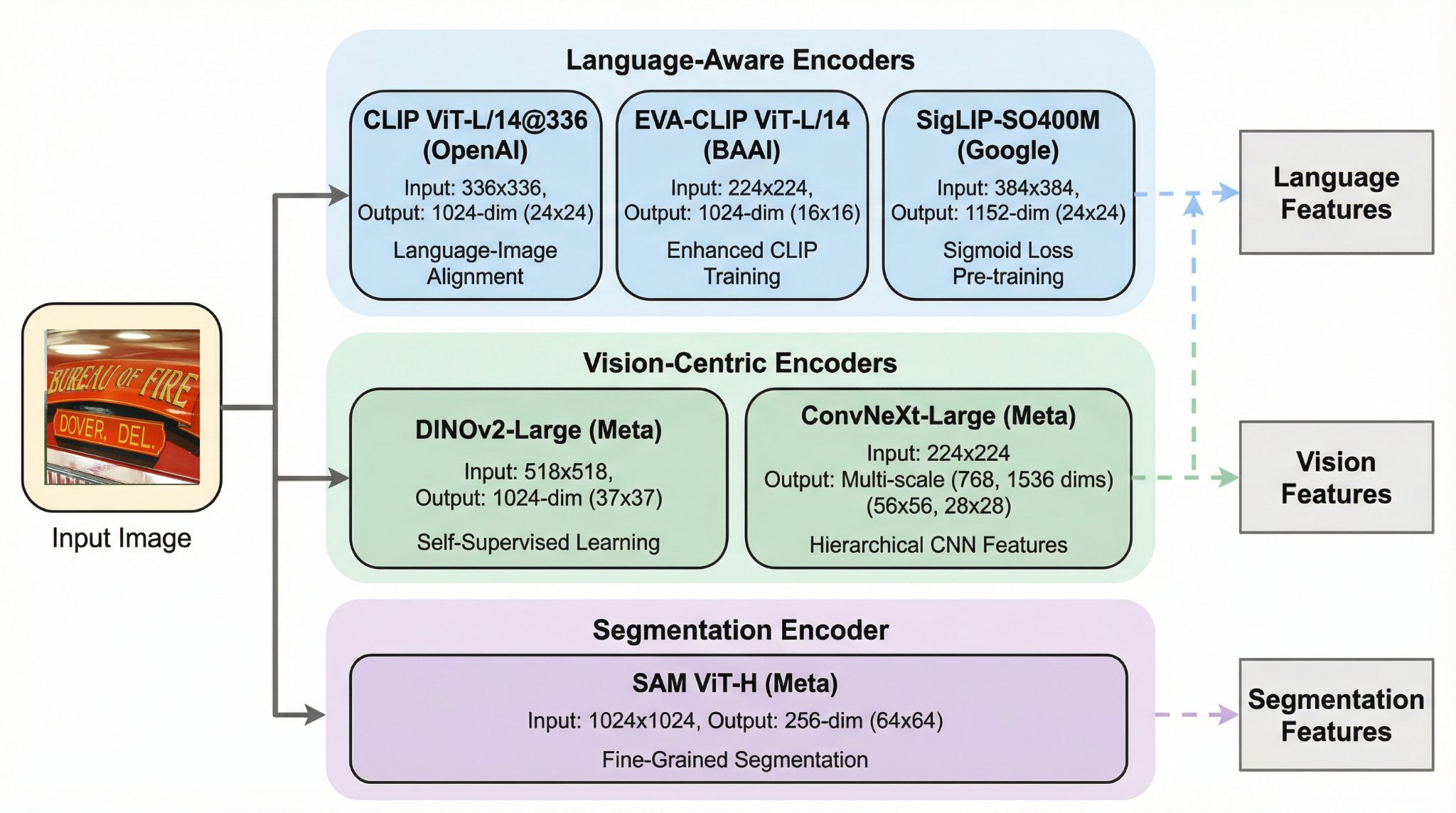}
\caption{The six frozen vision encoders spanning semantic (CLIP~\cite{radford2021clip}, EVA-CLIP~\cite{sun2024evaclip}, SigLIP~\cite{zhai2023siglip}), spatial (DINOv2~\cite{oquab2023dinov2}, SAM~\cite{kirillov2023sam}), and texture (ConvNeXt~\cite{liu2022convnext}) feature spectra.}
\label{fig:encoders}
\end{figure}

Single encoders trained on one objective cannot capture the wide variation in font, scale, orientation, and background clutter present in scene text. To address this, we use an ensemble of six frozen pre-trained encoders spanning language-image models, self-supervised transformers, segmentation architectures, and hierarchical CNNs. Each encoder $E_i$ produces features $\mathbf{f}_i \in \mathbb{R}^{C_i \times H_i \times W_i}$, and the ensemble output is the set $\{\mathbf{f}_1, \ldots, \mathbf{f}_6\}$. Figure~\ref{fig:encoders} summarises the configurations.

\begin{figure*}[!htb]
\centering
\includegraphics[width=0.6\textwidth]{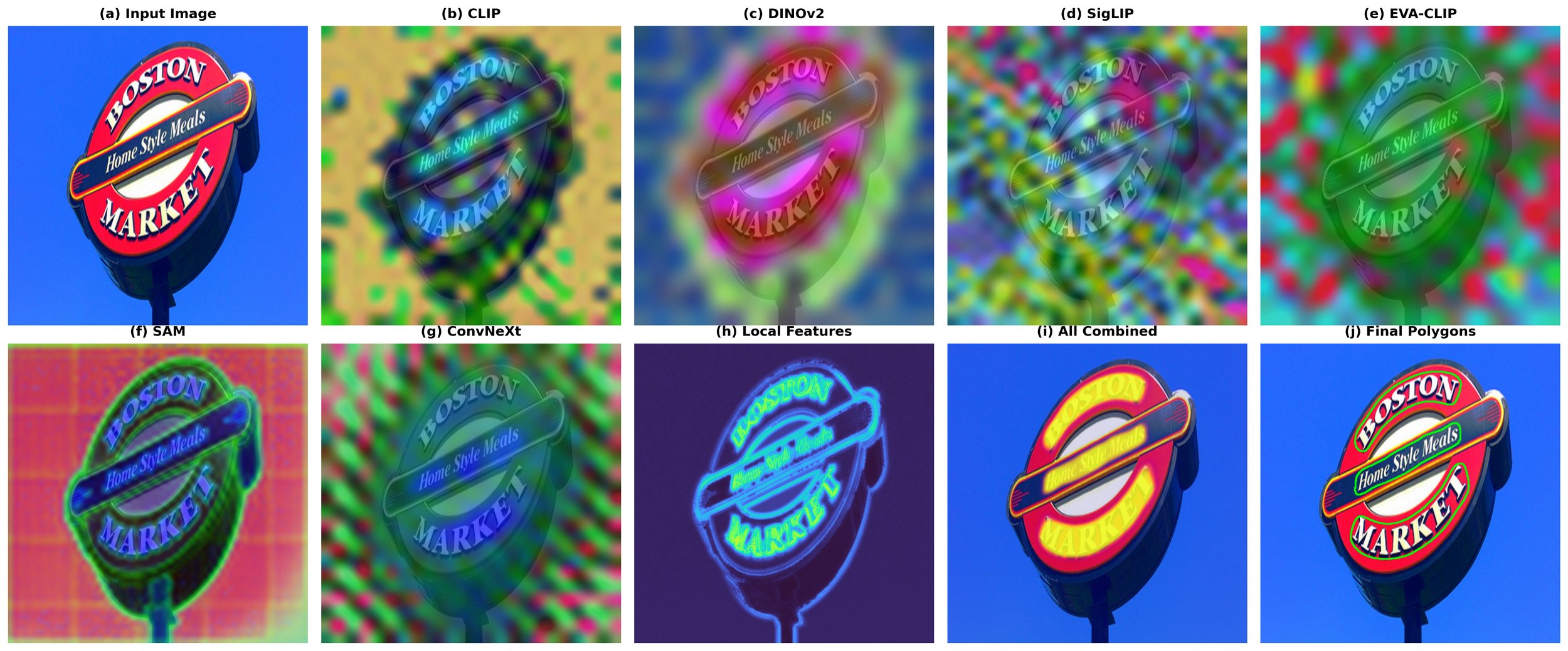}
\caption{Per-encoder PCA-RGB feature visualizations on a CTW1500 sample.
(b--g) show all six encoders' projected features: semantic patches (CLIP),
smooth instance blobs (DINOv2), language-aligned semantics (SigLIP, EVA-CLIP),
boundary-aware spatial structure (SAM), and fine texture (ConvNeXt).
(h) Handcrafted local features. (i) Full PSE confidence. (j) Final polygons.}

\label{fig:encoder_contributions}
\end{figure*}

The six encoders are selected to span three complementary feature spectra. CLIP, SigLIP, and EVA-CLIP provide \textit{semantic} features through language and image pretraining, capturing what visual content represents. DINOv2 and SAM provide \textit{spatial} features through self supervised and segmentation pretraining, capturing where text boundaries lie. ConvNeXt provides \textit{texture} features through its hierarchical CNN architecture, preserving local stroke patterns that transformer encoders tend to smooth out. All encoder outputs are projected to a common dimension and concatenated channel-wise. As shown in Table~\ref{tab:ablation_all}, progressively adding encoders from each spectrum consistently improves detection, confirming that each contributes meaningfully. Figure~\ref{fig:encoder_contributions} visualizes the per-encoder feature responses.

\vspace{-1.5em}
\subsubsection{Local Feature Extraction:}

\begin{figure}[!htb]
\centering
\includegraphics[width=0.62\linewidth]{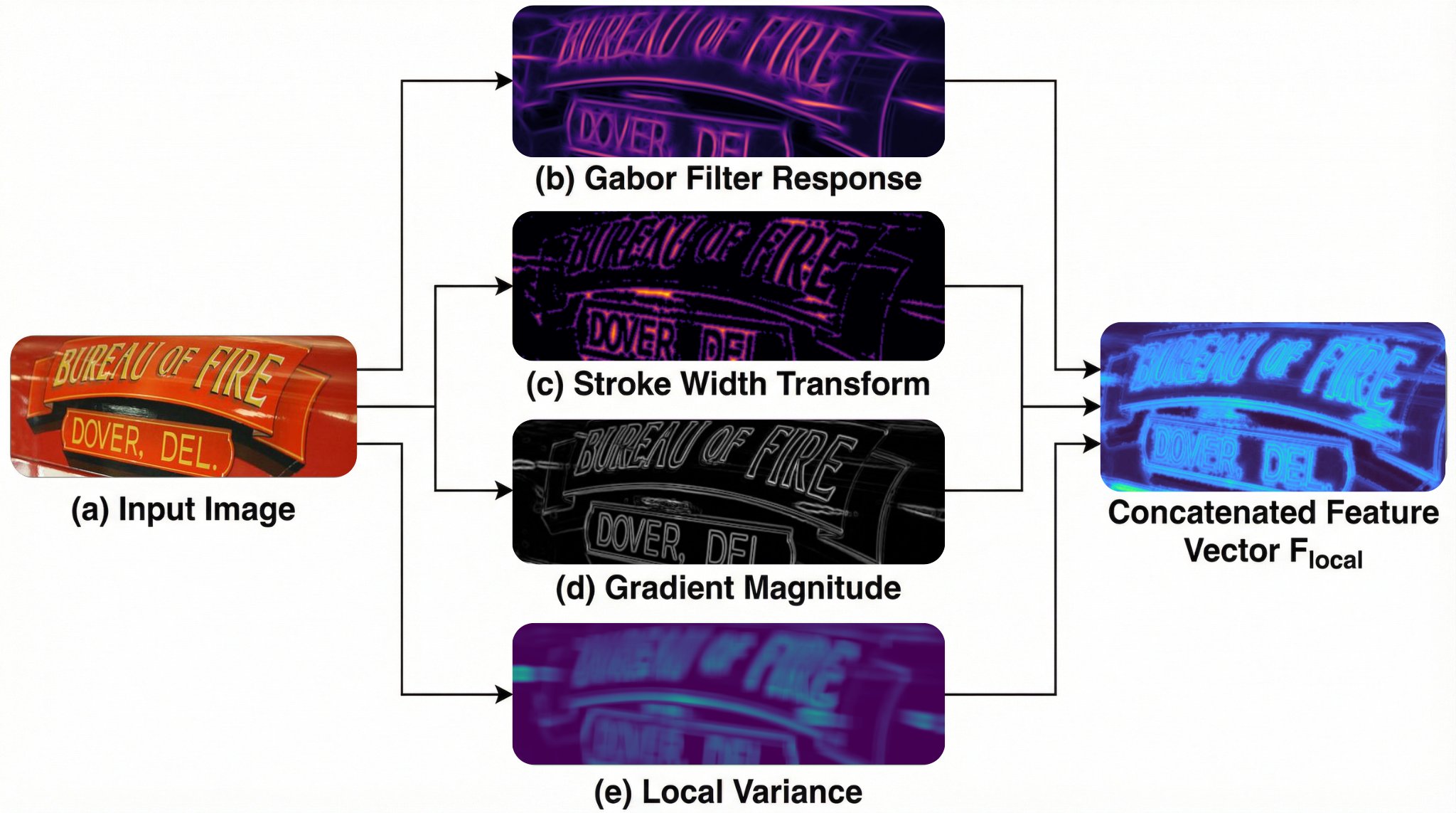}
\caption{Local feature channels: Gabor responses capture stroke texture at multiple orientations, while Sobel edges, stroke width, and gradient magnitude provide sub-pixel boundary cues for polygon refinement.}

\label{fig:local_features}
\end{figure}

To supply sub-pixel boundary cues that coarse-resolution encoders smooth out, we extract 16 handcrafted channels $\mathbf{f}_{local} \in \mathbb{R}^{16 \times 32 \times 32}$: Gabor responses at multiple orientations, Sobel edges, stroke-width estimates, CLAHE enhancement, local variance, edge density, and gradient magnitude. These channels are processed through a separate convolutional branch and concatenated with the global FPN output. Their primary role is to sharpen polygon vertices during PSE boundary expansion, improving crop quality for downstream recognition. Figure~\ref{fig:local_features} shows sample visualizations. These are processed through three convolutional blocks to 512 channels and concatenated with the FPN output to form $\mathbf{F}_{comb} \in \mathbb{R}^{1024 \times 32 \times 32}$.

\subsection{Multi-Scale Feature Pyramid Network}

Text instances span a wide range of sizes, so a fixed-resolution feature map cannot handle both small and large text well. We build a Multi-Scale FPN~\cite{lin2017fpn} that fuses encoder outputs at multiple resolutions. Each encoder output is projected to a common dimension $d=512$ via $1\!\times\!1$ convolutions, resized to three pyramid scales ($8^2$, $16^2$, $32^2$), and fused by concatenation followed by a $1\!\times\!1$ convolution at each scale:
\begin{equation}
\mathbf{F}_l = \text{Conv}_{1\times1}\!\Big(\text{Concat}\big[\text{Proj}_i(\mathbf{f}_i)\big]_{i=1}^{6}\Big), \quad l \in \{1,2,3\}
\end{equation}
where $\mathbf{F}_l$ is the fused feature map at pyramid level~$l$, $\text{Proj}_i$ projects encoder~$i$'s output to $d\!=\!512$ channels, and $\text{Conv}_{1\times1}$ reduces the concatenated channels back to~$d$.

\begin{figure*}[!htb]
\centering
\begin{subfigure}[b]{0.19\textwidth}
  \centering\includegraphics[height=2.3cm]{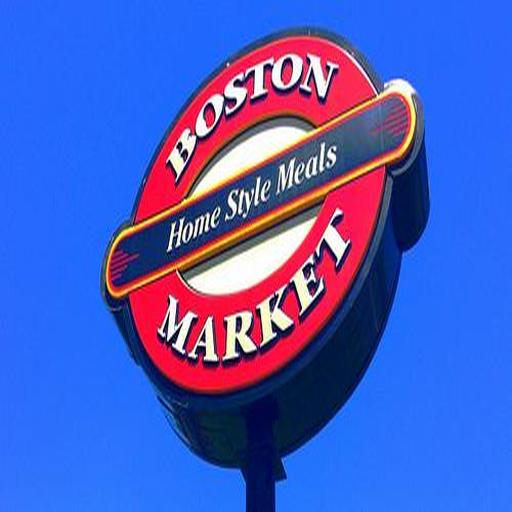}
  \caption{Input}
\end{subfigure}
\hfill
\begin{subfigure}[b]{0.19\textwidth}
  \centering\includegraphics[height=2.3cm]{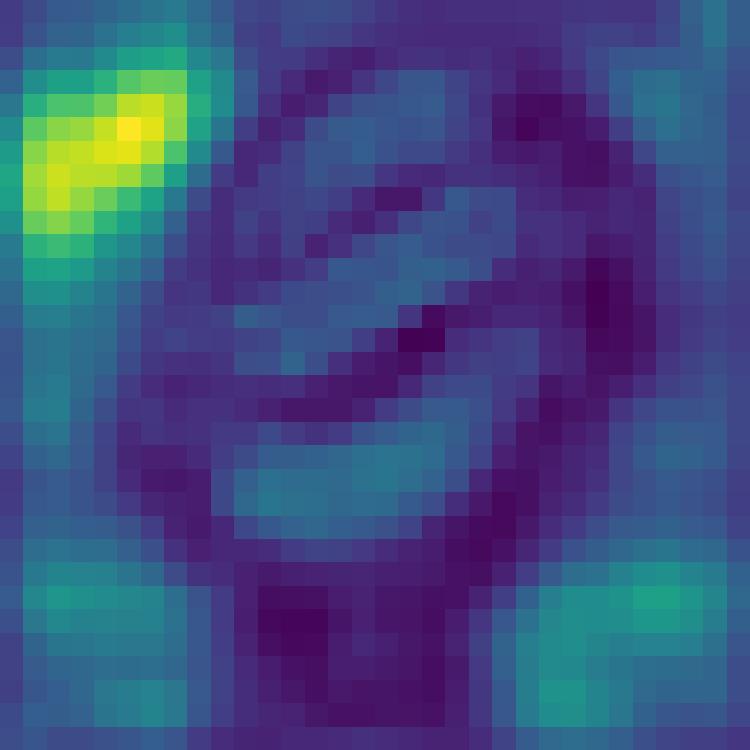}
  \caption{Features}
\end{subfigure}
\hfill
\begin{subfigure}[b]{0.19\textwidth}
  \centering\includegraphics[height=2.3cm]{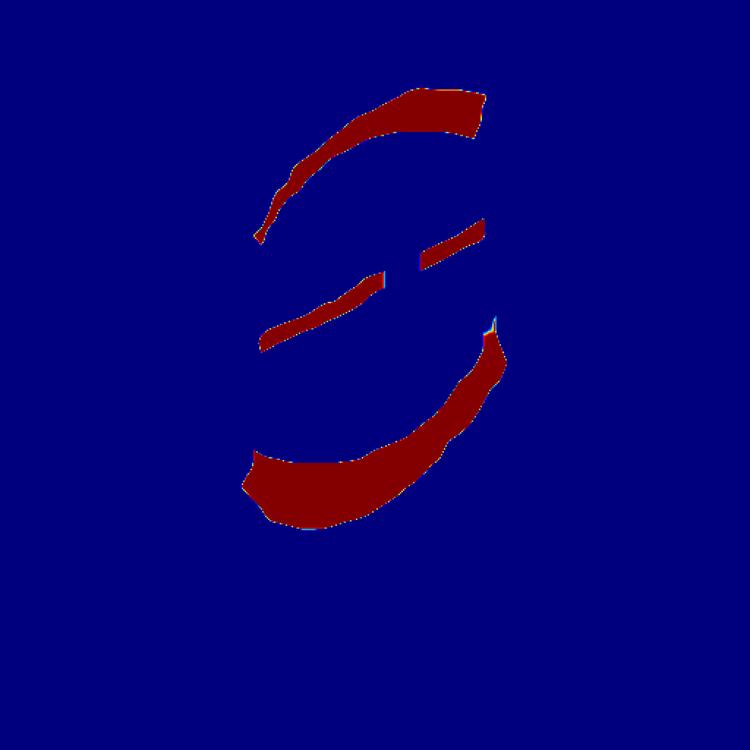}
  \caption{Kernel $S_1$}
\end{subfigure}
\hfill
\begin{subfigure}[b]{0.19\textwidth}
  \centering\includegraphics[height=2.3cm]{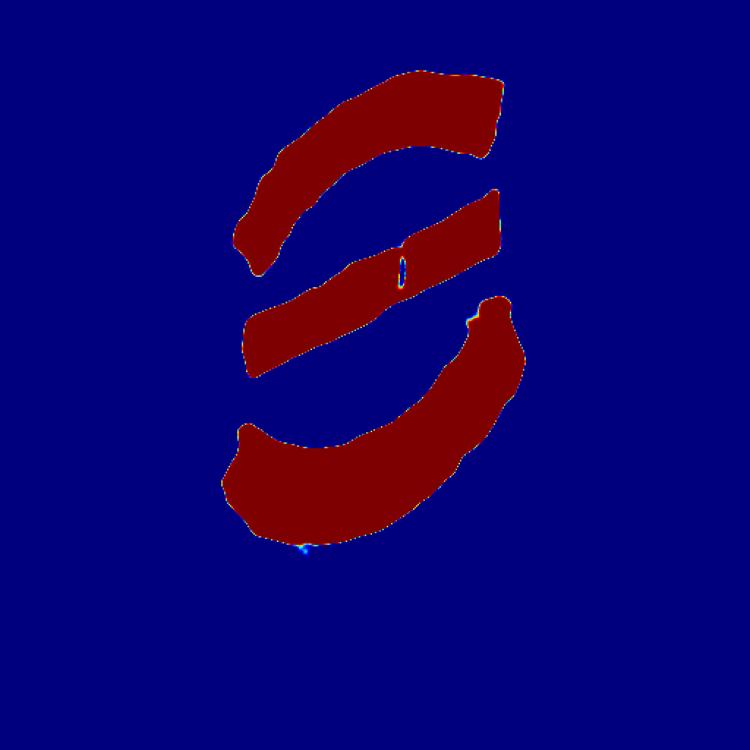}
  \caption{Kernel $S_6$}
\end{subfigure}
\hfill
\begin{subfigure}[b]{0.19\textwidth}
  \centering\includegraphics[height=2.3cm]{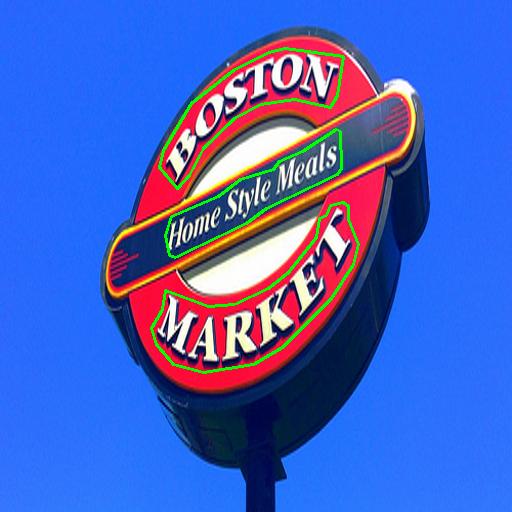}
  \caption{Polygons}
\end{subfigure}
\caption{Intermediate outputs of the detection pipeline on a sample from CTW1500. (a) Input image containing the curved sign. (b) Fused multi-encoder feature map showing strong activation on text regions. (c) Smallest PSE kernel $S_1$ highlighting text center seeds. (d) Largest kernel $S_6$ covering the full text extent. (e) Final polygon instances after progressive scale expansion.}
\label{fig:step_by_step}
\end{figure*}
\vspace{-2.1em}
\subsubsection{Deformable Attention~\cite{zhu2021deformable}} learns to sample only at task-relevant locations via learnable offsets:
\begin{equation}
\text{DeformAttn}(q, \{V_l\}) = \sum_{l=1}^{L} \sum_{p=1}^{P} A_{l,p} \cdot V_l(\mathbf{r} + \Delta\mathbf{p}_{l,p})
\end{equation}
where $\mathbf{r}$ is the reference point, $\Delta\mathbf{p}_{l,p}$ are learned offsets, and $A_{l,p}$ are attention weights. We set $L\!=\!3$ levels to match our three FPN pyramid scales, $P\!=\!8$ sampling points per level following the default configuration of Deformable DETR~\cite{zhu2021deformable}, and 8 attention heads to balance capacity with computational cost.

\subsection{Progressive Scale Expansion (PSE)}

Adjacent text instances often merge into a single region during segmentation. PSE addresses this by predicting $K\!=\!6$ kernel maps at increasing scales, where the smallest kernel captures only the text centre and subsequent kernels expand outward without merging neighbours.

The PSE decoder maps $\mathbf{F}_{comb}$ to $K$ kernel maps through an encoder convolution followed by four transposed convolutions (1024, 512, 256, 128, 64 channels), with a final head producing output $\mathbf{S} \in \mathbb{R}^{K \times 512 \times 512}$:
\begin{equation}
\mathbf{S} = \text{PSEDecoder}(\mathbf{F}_{comb}), \quad S_k(x,y) \in [0,1]
\end{equation}
where $\mathbf{F}_{comb}$ is the combined global and local feature map, $K\!=\!6$ is the number of scale kernels, and $S_k(x,y)$ is the text confidence at pixel $(x,y)$ for kernel~$k$.

The smallest kernel $S_1$ is thresholded at $\tau_s\!=\!0.4$ and labelled via connected components. Each subsequent kernel $S_k$ ($k\!=\!2,...,6$), thresholded at $\tau_k\!=\!0.5$, is expanded via breadth-first search while preserving instance boundaries. Contours below 50 pixels are discarded, and the remaining are simplified with Douglas-Peucker ($\epsilon = 0.002 \times \text{perimeter}$) to produce final polygons.  Figure~\ref{fig:step_by_step} illustrates these intermediate outputs on a CTW1500 sample.

\vspace{-2.3em}
\subsection{\rev{Decoupled Feature Interaction}}
\vspace{-0.1em}
A core challenge in arbitrarily shaped scene text spotting is achieving precise boundaries. In traditional single-backbone detectors, an inherent \textit{semantic-spatial conflict} exists: features optimized to confidently recognize text semantics tend to smooth over the sharp edge gradients required for tight, accurate localization. 

We address this by treating our Progressive Scale Expansion (PSE) process as a decoupled interaction between an expansion cue and a stopping cue. A polygon can be read, qualitatively, as the balance of a total energy in which the expansion force is kept apart from the stopping constraints:
\begin{equation}
\label{eq:energy}
E_{total} = E_{semantic}(\text{expansion}) + E_{spatial}(\text{stopping})
\end{equation}
where $E_{semantic}$ is the expansive energy from semantic encoders that identifies text cores, and $E_{spatial}$ is the stopping energy from spatial encoders and local features that anchors boundaries to physical edges. \rev{Eq.~\ref{eq:energy} is descriptive and is not a trained objective. We do not minimise any energy term during optimisation, and we add no extra loss or gradient for it. The equation only describes how the decoupled features shape the PSE expansion in our model, with the expansion (semantic) and stopping (spatial) cues coming from frozen feature spaces that stay orthogonal.}

Our architecture directly implements this decoupling. The semantic encoders (e.g., CLIP, SigLIP) provide the expansive internal energy ($E_{semantic}$), Simultaneously, the spatial encoders (e.g., SAM, DINOv2) and low-level handcrafted features provide the external stopping energy ($E_{spatial}$), anchoring the expansion to physical visual boundaries like character strokes and edge gradients. Because our foundation encoders are frozen and independently trained, they maintain orthogonal feature spaces during decoding. This prevents the homogenization of features that degrades boundaries in unified networks, allowing the PSE algorithm to halt expansion exactly at the high-frequency text edges.
\vspace{-1.5em}
\subsection{Vision-Language Model for Text Recognition}

Traditional OCR engines rely on fixed character dictionaries, limiting generalization to diverse scripts. VLMs jointly reason over visual and textual features without a predefined lexicon.

We select Qwen3-VL-8B-Instruct~\cite{qwen3technicalreport} as the recognition backbone for its strong scene-text visual grounding from large-scale image-text pretraining, which outperformed comparably-sized alternatives in preliminary recognition experiments on CTW1500 crops. Fine-tuning all 8B parameters would be expensive and prone to overfitting, so we apply Low-Rank Adaptation (LoRA)~\cite{hu2022lora}:
\begin{equation}
W' = W + BA, \quad B \in \mathbb{R}^{d \times r},\; A \in \mathbb{R}^{r \times k},\; r = 16 \ll \min(d,k)
\end{equation}
with $\alpha\!=\!32$ and dropout 0.05, applied to query, key, value, and output projections. This adapts ${\sim}0.2\%$ of parameters.

Text crops are extracted from polygon annotations with 5-pixel padding, resized to fit a $256\!\times\!256$ canvas while preserving aspect ratio, and padded with white background. The VLM is fine-tuned for 4 epochs with AdamW ($\text{lr}\!=\!10^{-4}$, cosine decay, 5\% warmup, batch size 16, bfloat16). The loss is selective cross-entropy on assistant response tokens only:
\begin{equation}
\mathcal{L}_{rec} = -\frac{1}{|\mathcal{A}|} \sum_{i \in \mathcal{A}} \log P(y_i \mid y_{<i}, \mathbf{x})
\end{equation}
where $\mathcal{A}$ is the set of assistant token positions, $y_i$ is the $i$-th target token, $y_{<i}$ are preceding tokens, and $\mathbf{x}$ is the input crop.
\vspace{-1em}
\subsection{End-to-End Training}

The detection and recognition stages are first trained independently and then connected through a three-phase schedule. The total loss is:
\begin{equation}
\mathcal{L} = \mathcal{L}_{det} + \lambda_{rec} \cdot \mathcal{L}_{rec}, \quad \mathcal{L}_{det} = \mathcal{L}_{BCE} + \mathcal{L}_{Dice}
\end{equation}
where $\mathcal{L}_{BCE}$ is binary cross-entropy, $\mathcal{L}_{Dice}$ is dice loss for segmentation, and $\lambda_{rec}$ controls the recognition loss weight.

The phased approach stabilises learning by allowing detection to converge before introducing the heavier VLM gradients. Phase~1 (epochs 1--2) trains detection only ($\lambda_{rec}\!=\!0$). Phase~2 (epochs 3--10) introduces sparse VLM supervision every 16 steps with $\lambda_{rec}\!=\!0.2$ on 1 crop per batch. Phase~3 (epochs 11--30) increases to every 8 steps with $\lambda_{rec}\!=\!0.4$ on 2 crops per batch. We use separate learning rates: $\text{lr}_{det}\!=\!10^{-4}$ and $\text{lr}_{vlm}\!=\!10^{-5}$, with effective batch size 16 (8 $\times$ 2 accumulation steps) and bfloat16 throughout.

\noindent\textbf{Inference:}
At inference, the input is resized to $512\!\times\!512$, passed through the encoder ensemble and FPN, and the PSE decoder predicts six kernel maps. The PSE algorithm separates instances into polygons, filtering those with area $<200$ pixels or aspect ratio outside $[0.1, 15]$. Each polygon crops the corresponding region from the original-resolution image, and the fine-tuned VLM generates the transcription.
\vspace{-0.5em}
\section{Experiments}
\vspace{-0.2em}
\subsection{Datasets and Evaluation Metrics}
\vspace{-0.2em}
We evaluate on three benchmarks. \textbf{CTW1500}~\cite{liu2019ctw1500} (1,000/500 train/test images, 14-point polygon annotations) targets curved text at the line level. \textbf{ICDAR 2015}~\cite{karatzas2015icdar} (1,000/500 images, word-level quadrilaterals) contains blurry, multi-oriented text from a wearable camera. \textbf{Total-Text}~\cite{chng2017totaltext} (1,255/300 images, polygon annotations) covers horizontal, multi-oriented, and curved word instances.

We report Precision, Recall, and F1 for detection (IoU $\geq$ 0.5), and end-to-end H-mean under None (lexicon-free), Full (full test-set lexicon), and Weak (ICDAR~2015 specific) settings.

\subsection{Text Detection Results}
\vspace{-0.3em}
Table~\ref{tab:detection} compares our detection stage against established methods. Our multi-encoder ensemble achieves 92.0\% F1 on CTW1500, 89.6\% on Total-Text, and 90.8\% on ICDAR~2015, trained exclusively on each benchmark's training split with no synthetic data. Across all three datasets, our method consistently achieves the highest recall (93.8\%, 96.4\%, 93.5\%), indicating that the multi-encoder ensemble rarely misses text instances regardless of shape or orientation. The high recall is a direct consequence of fusing complementary feature spectra: spatial encoders anchor boundary locations while semantic encoders ensure even low-contrast text is detected. However, precision on Total-Text (83.9\%) is noticeably lower than on the other benchmarks, because the high-sensitivity ensemble occasionally fires on textured background regions that resemble stroke patterns. This precision-recall trade-off reflects our design choice to favour recall, since missed detections cannot be recovered by the downstream VLM, whereas false positives are discarded when the recognizer produces low-confidence or empty text outputs.

\rev{\noindent\textbf{Fairness of comparison.} None of the baselines in Tables~\ref{tab:detection} and~\ref{tab:spotting} is trained from scratch. They all start from ImageNet-pretrained backbones, so the comparison is really about which pretrained representation a method uses. The methods marked $*$ go further and use large-scale synthetic data together with multi-dataset supervision, which we do not use at all. We train MEVL-STP only on each benchmark's own split (1{,}000/1{,}255/1{,}000 images), and it still surpasses these baselines on CTW1500.}

\rev{\noindent\textbf{Precision--recall sensitivity and false-positive audit.} The lower Total-Text precision comes from our recall-favouring default seed threshold $\tau_s\!=\!0.4$; a higher $\tau_s$ trades recall for precision while F1 stays almost flat (Table~\ref{tab:tau_sweep}). Sweeping $\tau_s$ from 0.3 to 0.7 without retraining, Total-Text precision rises from 83.9\% to 87.3\% for a 3.1-point recall drop, and CTW1500 behaves the same with F1 near 92\%. A false-positive audit on CTW1500 is also reassuring: of 974 false positives, only 0.9\% come from texture, while 72.6\% are correct text localised at IoU 0.3 to 0.5 (boundary slack, not hallucination), and the recogniser drops a further 5.8\% by returning empty text.}

\begin{table}[!htb]
\centering
\caption{\rev{Seed-threshold $\tau_s$ sensitivity on CTW1500 and Total-Text, with no retraining. A higher $\tau_s$ raises precision while F1 stays almost flat.}}
\label{tab:tau_sweep}
{\small
\begin{tabular}{llccccc}
\toprule
& $\tau_s$ & 0.30 & 0.40 & 0.50 & 0.60 & 0.70 \\
\midrule
\multirow{3}{*}{CTW1500} & P & 87.67 & 88.99 & 90.19 & 91.32 & \textbf{92.39} \\
& R & \textbf{95.39} & 94.57 & 93.69 & 92.71 & 91.59 \\
& F1 & 91.37 & 91.69 & 91.91 & \textbf{92.01} & 91.99 \\
\midrule
\multirow{3}{*}{Total-Text} & P & 82.58 & 83.90 & 85.18 & 86.30 & \textbf{87.30} \\
& R & \textbf{97.20} & 96.40 & 95.30 & 94.40 & 93.30 \\
& F1 & 89.30 & 89.72 & 89.96 & 90.17 & \textbf{90.20} \\
\bottomrule
\end{tabular}}
\end{table}

\vspace{-2em}
\begin{table*}[!htb]
  \centering
  \caption{Comparison of text detection performance on all three datasets: CTW1500, Total-Text, and ICDAR 2015. * denotes methods trained with synthetic data. \textbf{Bold} and \underline{underline} indicate the best and second-best performance, respectively.}
  \vspace{2mm}
  \label{tab:detection}
  \resizebox{\textwidth}{!}{%
  \begin{tabular}{lllc|ccc|ccc|ccc}
    \toprule
    \multirow{2}{*}{\textbf{Type}} & \multirow{2}{*}{\textbf{Method}} & \multirow{2}{*}{\textbf{Venue}} & \multirow{2}{*}{\textbf{Year}}
    & \multicolumn{3}{c|}{\textbf{CTW1500}}
    & \multicolumn{3}{c|}{\textbf{Total-Text}}
    & \multicolumn{3}{c}{\textbf{ICDAR 2015}} \\
    & & & & P & R & F1 & P & R & F1 & P & R & F1 \\
    \midrule
    \multirow{9}{*}{Seg.}
    & TextBPN~\cite{zhang2020textbpn} & ICCV & 2021 & 88.3 & 84.7 & 86.5 & 92.4 & 87.9 & \underline{90.1} & -- & -- & -- \\
    & FSG~\cite{tang2022fsg} & CVPR & 2022 & 88.1 & 82.4 & 85.2 & 90.7 & 85.7 & 88.1 & 90.9 & 87.3 & 89.1 \\
    & CBNet~\cite{zhao2024cbnet} & IJCV & 2024 & 89.0 & 81.9 & 85.3 & 90.1 & 82.5 & 86.1 & -- & -- & -- \\
    & ABCNet v2*~\cite{liu2021abcnetv2} & TPAMI & 2022 & 85.6 & 83.8 & 84.7 & 89.2 & 84.1 & 87.0 & 90.4 & 86.0 & 88.1 \\
    & IAST*~\cite{zhang2024iast} & TIP & 2024 & 89.2 & 84.8 & 86.9 & \textbf{94.7} & 85.2 & 89.7 & 92.5 & 86.6 & 89.5 \\
    \midrule
    \multirow{7}{*}{Reg.}
    & DPText-DETR~\cite{ye2023dptext} & AAAI & 2023 & 91.7 & 86.2 & 88.8 & 91.8 & 86.4 & 89.0 & -- & -- & -- \\
    & DeepSolo*~\cite{ye2023deepsolo} & CVPR & 2023 & \textbf{93.2} & 85.0 & 88.9 & \underline{93.1} & 82.1 & 87.3 & \underline{92.8} & 87.4 & 90.0 \\
    & LRANet++~\cite{su2025lranet} & arXiv & 2026 & 89.1 & 85.3 & 87.2 & 91.1 & 85.2 & 88.1 & 89.6 & 86.7 & 88.1 \\
    & LRANet++*~\cite{su2025lranet} & arXiv & 2026 & \underline{92.8} & \underline{88.1} & \underline{90.3} & 92.6 & \underline{89.1} & \textbf{90.8} & \textbf{93.9} & 88.0 & \textbf{90.9} \\
    & OmniParser*~\cite{yu2025omniparser} & TPAMI & 2026 & 87.9 & 87.6 & 87.8 & 88.4 & 88.6 & 88.5 & 90.3 & \underline{91.0} & 90.7 \\
    \midrule
    & \textbf{MEVL-STP} & -- & 2026 & 90.3 & \textbf{93.8} & \textbf{92.0} & 83.9 & \textbf{96.4} & 89.6 & 88.2 & \textbf{93.5} & \underline{90.8} \\
    \bottomrule
  \end{tabular}%
  }
\end{table*}

\subsection{End-to-End Text Spotting Results}

Table~\ref{tab:spotting} compares our full pipeline against recent methods. Methods marked~* use large-scale synthetic and multi-dataset training. Despite training exclusively on each benchmark's training split, we achieve a Full H-mean of \textbf{85.9} on CTW1500, surpassing the prior best (LRANet++, 85.2) by +0.7 points, and set a new state of the art on CTW1500 None at \textbf{75.2}, a +4.5 margin over the previous best without any synthetic data. On Total-Text we achieve 81.9/85.6 (None/Full) and on ICDAR~2015 81.7 (Weak), though the remaining gap relative to synthetic-data methods on these benchmarks reflects the advantage of large-scale pretraining on word-level instances, which our method forgoes entirely.

\begin{table*}[!htb]
  \centering
  \caption{End-to-end text spotting results (H-mean \%). ``None'': lexicon-free evaluation; ``Full'': full test-set lexicon; ``W'': weak lexicon for ICDAR~2015 (all words in test set, equivalent to Full). Methods marked~* use large-scale synthetic and multi-dataset external training data. \textbf{Bold}: best; \underline{underline}: second best.}
  \vspace{2mm}
  \label{tab:spotting}
  \resizebox{\textwidth}{!}{%
  \begin{tabular}{llccccccc}
    \toprule
    & & & \multicolumn{2}{c}{\textbf{CTW1500}} & \multicolumn{2}{c}{\textbf{Total-Text}} & \multicolumn{1}{c}{\textbf{ICDAR 2015}} \\
    \cmidrule(lr){4-5}\cmidrule(lr){6-7}\cmidrule(lr){8-8}
    \textbf{Method} & \textbf{Venue} & \textbf{Year} & None & Full & None & Full & W \\
    \midrule
    TESTR*~\cite{zhang2022testr}                      & CVPR & 2022 & 56.0 & 81.5 & 73.3 & 83.9 & 79.4 \\
    TPSNet*~\cite{zheng2024tpsnet}                    & ACM MM & 2022 & 59.7 & 79.2 & 76.1 & 82.3 & --   \\
    ABINet++*~\cite{fang2023abinetpp}                 & TPAMI & 2023 & 60.2 & 80.3 & 77.6 & 84.5 & 80.4 \\
    DeepSolo*~\cite{ye2023deepsolo}                   & CVPR & 2023 & 64.2 & 81.4 & 82.5 & 88.7 & 83.5 \\
    ESTextSpotter*~\cite{huang2023estextspotter}       & ICCV & 2023 & 64.9 & 83.9 & 80.8 & 87.1 & 83.0 \\
    IAST*~\cite{zhang2024iast}                        & TIP & 2024 & 62.4 & 82.9 & 71.9 & 83.5 & 80.0 \\
    UNITS*~\cite{chen2024units}                       & CVPR & 2023 & 66.4 & 82.3 & 78.7 & 86.0 & --   \\
    OmniParser*~\cite{yu2025omniparser}               & TPAMI & 2026 & 66.8 & 85.1 & \textbf{84.0} & \underline{88.9} & \textbf{84.5} \\
    LSGSpotter*~\cite{wan2024lsgspotter}              & AAAI & 2025 & 68.9 & 84.4 & 81.5 & 87.3 & --   \\
    LRANet++*~\cite{su2025lranet}                     & arXiv & 2026 & \underline{70.7} & \underline{85.2} & \underline{84.6} & \textbf{89.7} & \underline{84.0} \\
    \midrule
    \textbf{Ours}                                     & -- & 2026 & \textbf{75.2} & \textbf{85.9} & 81.9 & 85.6 & 81.7 \\
    \bottomrule
  \end{tabular}%
  }
\end{table*}

\subsection{Ablation Study}

We conduct experiments on CTW1500 (Table \ref{tab:ablation_all}) to validate our design choices. 

\begin{table}[!htb]
\centering
\caption{Ablation study on CTW1500. Dashes indicate configurations where the metric is not applicable.}
\label{tab:ablation_all}
\resizebox{\columnwidth}{!}{
\begin{tabular}{lcccc}
\toprule
\multirow{2}{*}{\textbf{Configuration}} 
& \multicolumn{2}{c}{\textbf{E2E Spotting}} 
& \multicolumn{2}{c}{\textbf{Recognition Only}} \\
\cmidrule(lr){2-3} \cmidrule(lr){4-5}
& \textbf{Det. F1} & \textbf{H-mean} 
& \textbf{Raw Acc.} & \textbf{Full Acc.} \\
\midrule

\multicolumn{5}{l}{\textit{Feature Spectrum Ablation}} \\
Semantic Only (CLIP, SigLIP, EVA) & 0.32 & -- & -- & -- \\
Spatial Only (DINOv2, SAM) & 72.88 & -- & -- & -- \\
Semantic + Texture (no Spatial) & 64.77 & -- & -- & -- \\
Spatial + Texture (no Semantic) & 90.32 & -- & -- & -- \\
Semantic + Spatial (no Texture) & 90.63 & -- & -- & -- \\
All Three Spectra (No Local feat.) & 91.92 & 75.20 & -- & -- \\
All Three Spectra & \textbf{91.99} & \textbf{75.20} & -- & -- \\

\addlinespace
\multicolumn{5}{l}{\textit{Training and Generalization}} \\
Joint Training & 88.05 & 56.85 & -- & -- \\
Cross-Dataset Eval & 91.46 & 79.60 & -- & -- \\

\addlinespace
\multicolumn{5}{l}{\textit{Recognition under Different Cropping Strategies}} \\
GT Polygons & -- & -- & 82.25 & 94.23 \\
Full Image (No Det. Stage) & -- & -- & 77.56 & -- \\
Rect. Bounding Box & -- & 51.16 & 73.65 & 84.40 \\

\bottomrule
\end{tabular}
}
\end{table}

\textbf{Feature Spectrum Interdependence:} No single feature type is self-sufficient. Semantic encoders alone reach only 0.32\% F1: they recognise \textit{what} is text but give diffuse activations that collapse without spatial grounding. Spatial encoders alone reach 72.88\% F1, giving boundary structure but weak content discrimination. Two spectra recover most performance (90.32--90.63\%), and the third lifts it to 91.92\%, so each spectrum adds complementary information.

\rev{\noindent\textbf{Per-encoder contribution.} To check that the gain comes from the architecture and not from raw capacity, we mask one encoder at inference without retraining (Table~\ref{tab:perenc}). Dropping a spatial encoder hurts the most (SAM $-7.92$, DINOv2 $-4.05$ F1), while the three semantic and texture encoders cost less than 1 F1 each, since CLIP and EVA-CLIP cover for each other in the semantic spectrum. A redundant ensemble would lose accuracy slowly under any removal; ours instead breaks down on the spatial encoders, as expected when the encoders are complementary. Linear CKA on 100 CTW1500 images supports this: across all 15 pairs the similarity stays near zero (mean 0.002, at most 0.023 for CLIP/EVA-CLIP), far below the roughly 0.4 typical of heads that share a backbone.}

\begin{table}[!htb]
\centering
\caption{\rev{Per-encoder ablation. Change in CTW1500 detection metrics when one encoder is masked at inference, with no retraining. The spatial encoders (SAM, DINOv2) contribute the most.}}
\label{tab:perenc}
{\footnotesize\setlength{\tabcolsep}{4.5pt}\renewcommand{\arraystretch}{0.9}
\begin{tabular}{lcccccc}
\toprule
Removed & CLIP & DINOv2 & SigLIP & EVA-CLIP & SAM & ConvNeXt \\
\midrule
$\Delta$F1 & $+0.06$ & $-4.05$ & $-0.80$ & $+0.04$ & $-7.92$ & $-0.89$ \\
$\Delta$P  & $+0.66$ & $+1.92$ & $-1.81$ & $+1.13$ & $+3.57$ & $+3.21$ \\
$\Delta$R  & $-0.61$ & $-10.05$ & $+0.40$ & $-1.18$ & $-18.21$ & $-5.19$ \\
\bottomrule
\end{tabular}}
\end{table}

\textbf{Training Strategy:} Joint training collapses end-to-end H-mean to 56.85\% despite maintaining 88.05\% detection F1, because VLM gradients destabilise the detection stage before polygon quality converges, confirming that phased training is essential. \rev{The conflict here is between two objectives. The detector pushes the polygons toward the ground-truth text region, while the recogniser prefers polygons whose crops are easy to read. When the two are trained together while the polygons are still noisy, these objectives pull in different directions. Phasing avoids this by attaching the recogniser only after the polygons have converged. On the same weights this recovers an 85.86\% H-mean, a gap of 29 points that turns on a single configuration switch, and it is exactly the error-decoupling benefit that our two-stage design has over tightly coupled end-to-end spotters.} Cross-dataset evaluation (train on CTW1500, test on Total-Text) yields 91.46\% detection F1 and 79.60\% H-mean, demonstrating that the frozen encoder ensemble generalises across datasets without domain-specific fine-tuning. Feeding the entire uncropped image directly to the VLM yields only 77.56\% accuracy, as it merges adjacent text. Cropping with loose rectangular bounding boxes achieves only 73.65\% accuracy because of background noise. Our exact polygon cropping removes this noise entirely, boosting accuracy to 82.25\%.

\rev{\noindent\textbf{Computational Cost.} MEVL-STP trains only 27.1\,M parameters (16\,M of LoRA on Qwen3-VL, 11\,M in the FPN, PSE decoder, and channel attention), fewer than DeepSolo's 42.5\,M; the six frozen encoders add 2.2\,B parameters at inference. The cost sits in one place and is tunable: at $1024^2$ input the detector uses 3601 GFLOPs, of which SAM-ViT-H alone is 2742 (76\%), and throughput is 3.18/1.13 FPS (detection/full) on an H100. Replacing SAM-ViT-H with SAM-ViT-L makes that branch about $4\times$ cheaper, and since SAM adds at most 7.92 F1 (Table~\ref{tab:perenc}), this trades accuracy for speed directly. Distilling the frozen ensemble into one student backbone, in the line of recent multi-teacher distillation, could remove this overhead while keeping the feature coverage; we leave it to future work.}

\vspace{-1.2em}
\subsection{Qualitative Results}
\begin{figure}[H]
    \centering
    \begin{subfigure}[b]{0.25\textwidth}
        \includegraphics[width=\textwidth]{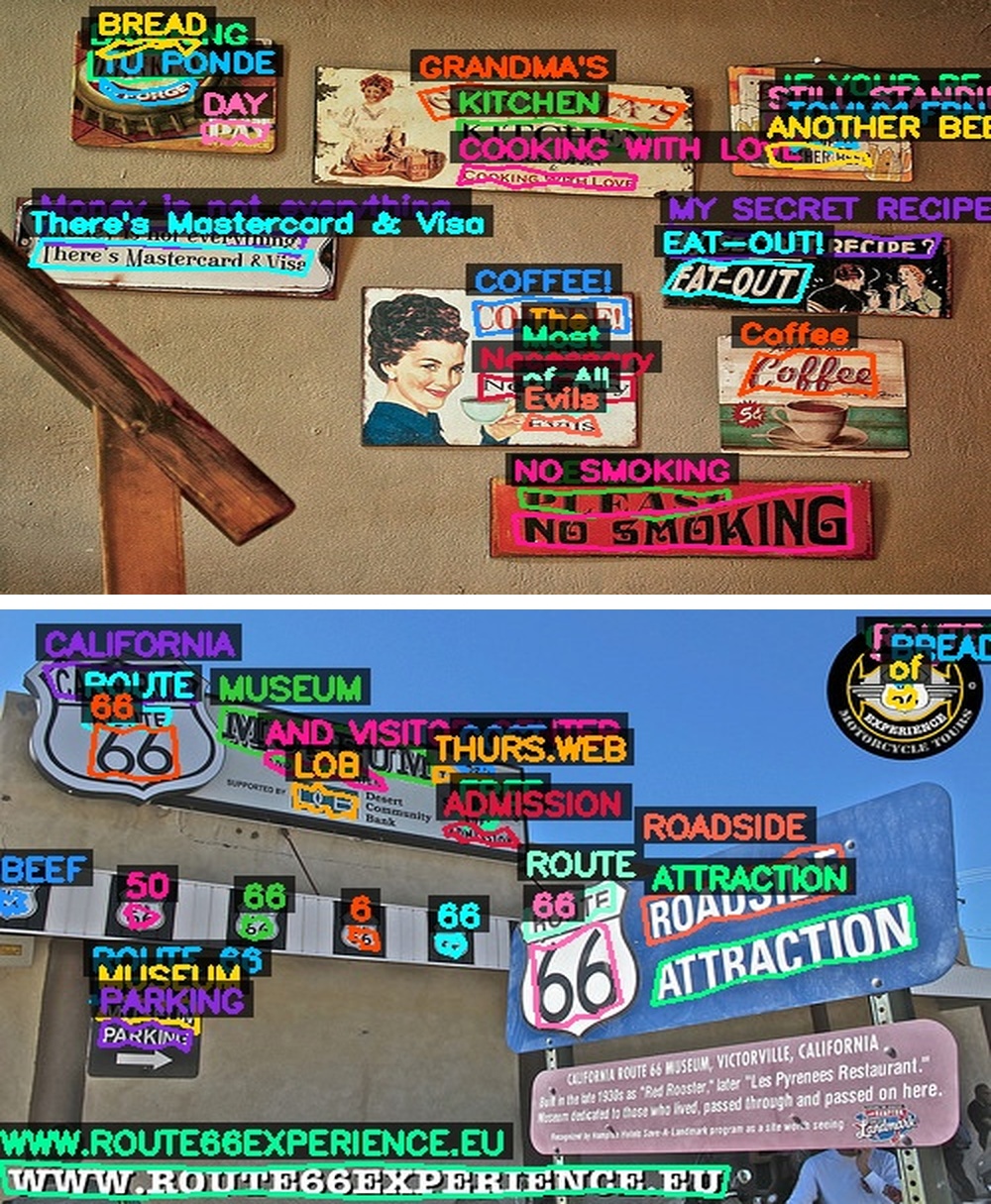}
        \caption{CTW1500}
    \end{subfigure}
    \hfill
    \begin{subfigure}[b]{0.25\textwidth}
        \includegraphics[width=\textwidth]{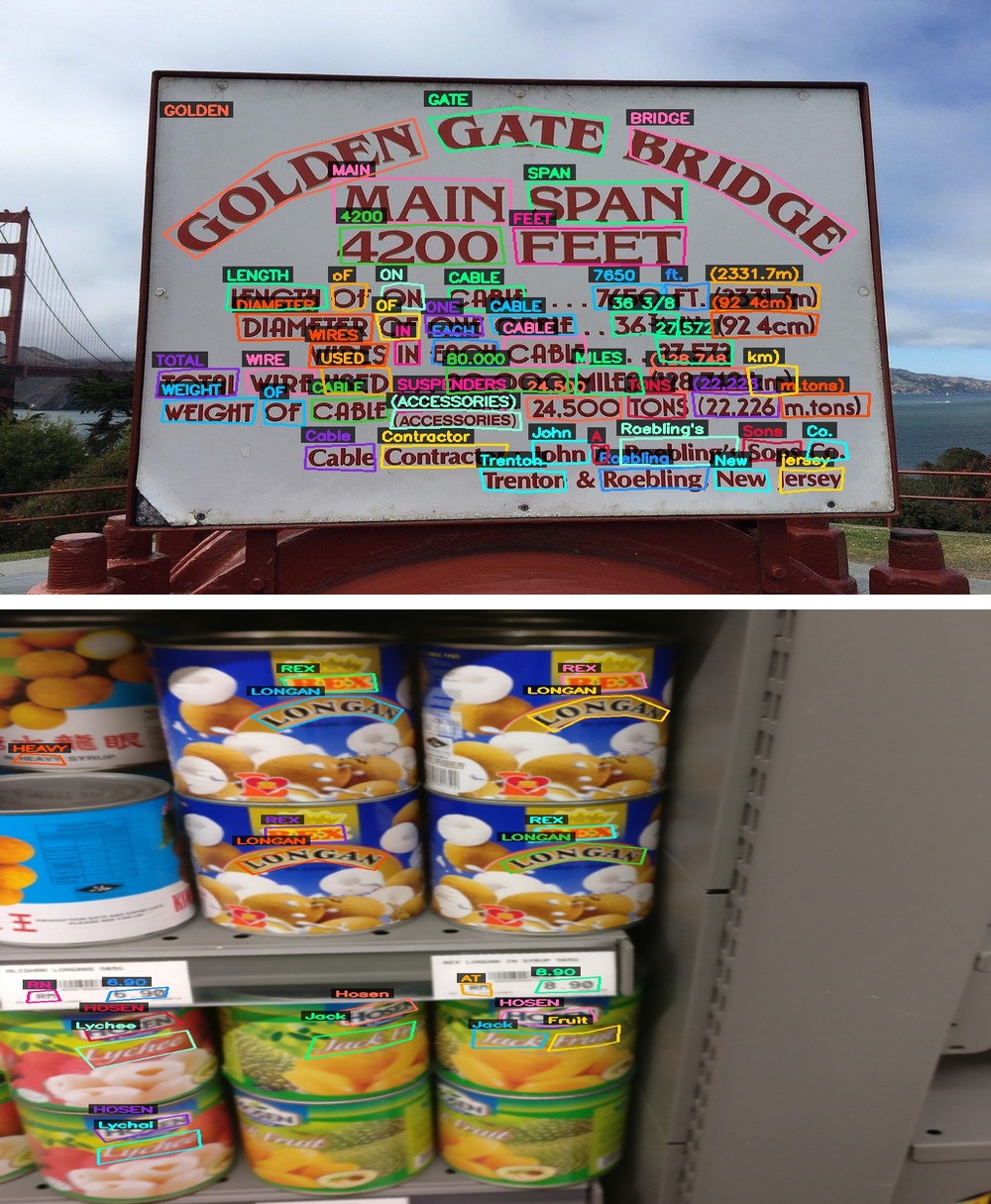}
        \caption{Total-Text}
    \end{subfigure}
    \hfill
    \begin{subfigure}[b]{0.25\textwidth}
        \includegraphics[width=\textwidth]{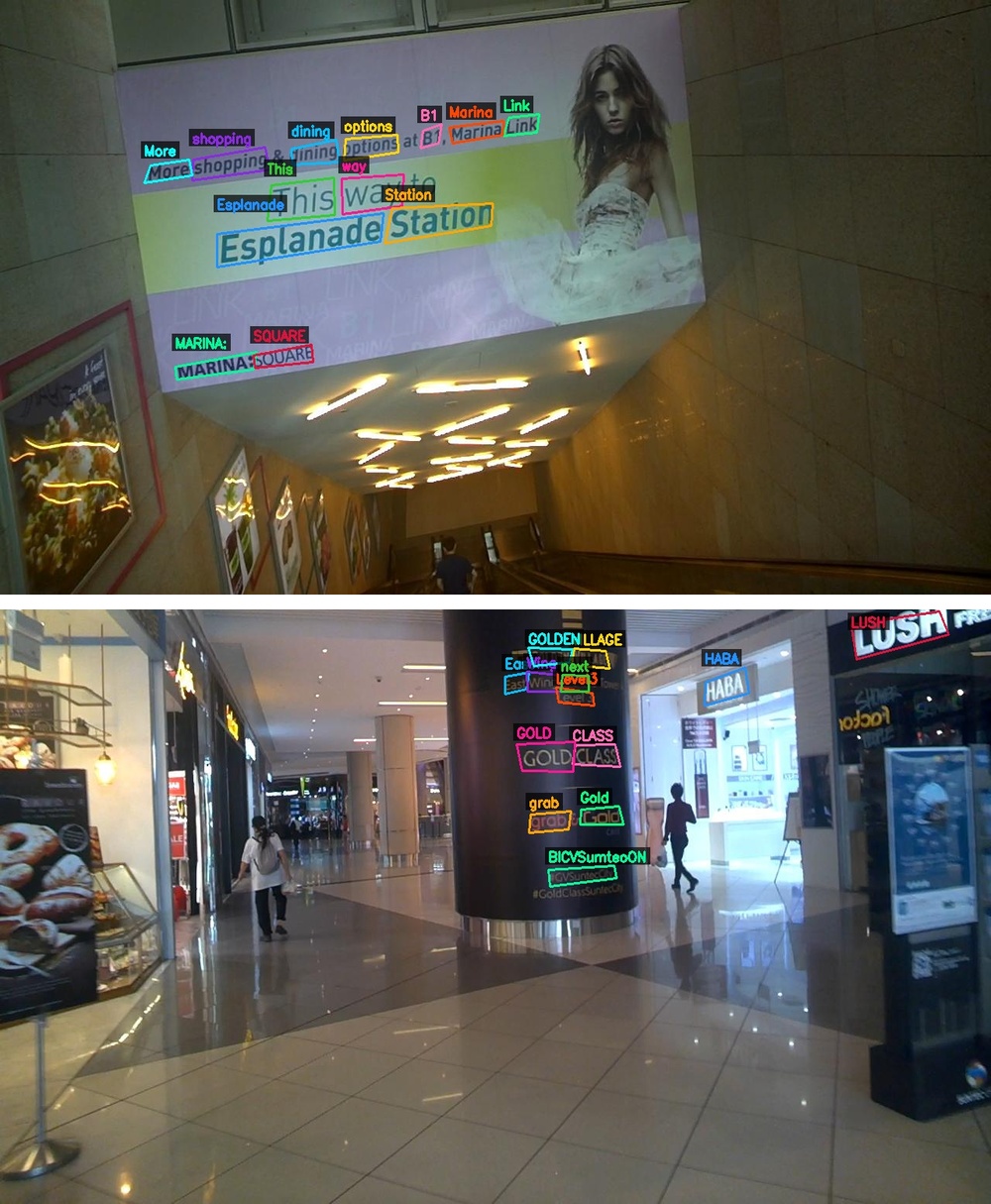}
        \caption{ICDAR 2015}
    \end{subfigure}
    \caption{Qualitative results on standard benchmarks demonstrating strong generalization across various scene text challenges, including extreme curvature (CTW1500), arbitrary orientations (Total-Text), and cluttered environmental text (ICDAR 2015).}
    \label{fig:qualitative}
\end{figure}

\vspace{-0.2em}
Figure~\ref{fig:qualitative} presents end-to-end recognition results across all three benchmarks. Our method handles curved text on signs (CTW1500), dense multi-oriented layouts (ICDAR 2015), and mixed horizontal and curved word instances (Total-Text).

\section{Conclusion}
\vspace{-0.2em}
We presented MEVL-STP, a decoupled text spotting framework that fuses six frozen foundation encoders across semantic, spatial, and texture spectra for detection and applies LoRA-adapted VLM recognition on tight polygon crops. The three spectra are codependent: no single type suffices, yet together they reach 91.99\% detection F1 on CTW1500 without synthetic data and a new state of the art of 85.86\% Full H-mean, while staying competitive on Total-Text and ICDAR 2015.

Limitations. The high-sensitivity ensemble occasionally fires on textured backgrounds, which lowers Total-Text precision; adaptive threshold calibration or a light verification stage would help. Recognition is also sequential per crop, so latency grows with the number of instances, which the distillation and batched inference noted above could reduce.

\section*{\rev{Acknowledgments}}
\rev{The authors thank the anonymous reviewers and the Area Chairs, whose feedback helped us improve the paper. This research received no specific grant from any funding agency in the public, commercial, or not-for-profit sectors.}

%
%
%
\bibliographystyle{unsrt}
\bibliography{references}

\end{document}